\documentclass[letterpaper]{article} 
\usepackage{aaai2027}  
\usepackage[hyphens]{url}  
\usepackage{graphicx} 
\usepackage{natbib}  
\usepackage{caption} 
\usepackage{algorithm}
\usepackage{algorithmic}
\usepackage{amssymb}
\usepackage{multirow}
\usepackage{placeins}

\usepackage{newfloat}
\usepackage{listings}
\DeclareCaptionStyle{ruled}{labelfont=normalfont,labelsep=colon,strut=off} 
\floatstyle{ruled}
\newfloat{listing}{tb}{lst}{}
\floatname{listing}{Listing}

\usepackage{booktabs}

\usepackage{amsmath}
\nocopyright

\title{FeedbackTrack: Visual-Cortex-Inspired Cross-Frame Feedback for Transformer Tracking}

\author{
    Yueyang Cang\textsuperscript{\rm 1},
    Xiaoteng Zhang\textsuperscript{\rm 1},
    Zhiyuan Ning\textsuperscript{\rm 1},
    Yuchen He\textsuperscript{\rm 1},
    Li Shi\textsuperscript{\rm 1}
}

\affiliations{
    \textsuperscript{\rm 1}Tsinghua University\\
    Beijing, China
}

\begin{document}

\maketitle

\begin{abstract}
Visual object tracking requires the continuous integration of target
information across frames, yet most Transformer trackers still rely on
predominantly feed-forward visual feature extraction.
Existing temporal mechanisms commonly update templates, prompts, queries,
or prediction states, while intermediate representations from previous
frames rarely modulate corresponding stages of current-frame processing.
We propose \textbf{FeedbackTrack}, a visual-cortex-inspired framework that
introduces sparse, group-level layer-aligned cross-frame feedback into
pretrained Transformer trackers.
Previous-frame outputs from selected Transformer groups are detached,
cached, and returned to the corresponding groups in the current frame.
FeedbackTrack contains two complementary pathways.
Query Feedback transforms historical search-token states into low-rank
query biases and applies per-sample, per-head, and per-token RMS alignment
to control their magnitude.
Gate Feedback uses pooled historical context to provide bounded
channel-group modulation of the projected attention output.
The proposed mechanism retains the original tracking pipeline and requires
only a fixed-size one-frame cache.
Across SPMTrack and ARTrackV2, FeedbackTrack consistently improves five
backbone configurations on LaSOT and GOT-10k.
It improves GOT-10k AO by 2.3--4.1 points and LaSOT AUC by 1.1--1.8
points, reaching 83.4 AO and 79.1 AUC with SPMTrack-G, while adding less
than 1\% parameters in all configurations.
Controlled comparisons further show that cross-frame feedback outperforms
same-frame modulation by 1.8--3.2 AO points, confirming that the gains
primarily arise from recurrent historical information rather than
additional modulation capacity.
Analysis of the learned feedback scales reveals a non-uniform
depth-dependent organization, demonstrating that recurrent cross-frame
feedback provides an effective and scalable mechanism for Transformer
tracking.
\end{abstract}

\begin{figure}[t]
    \centering
    \includegraphics[width=\columnwidth]{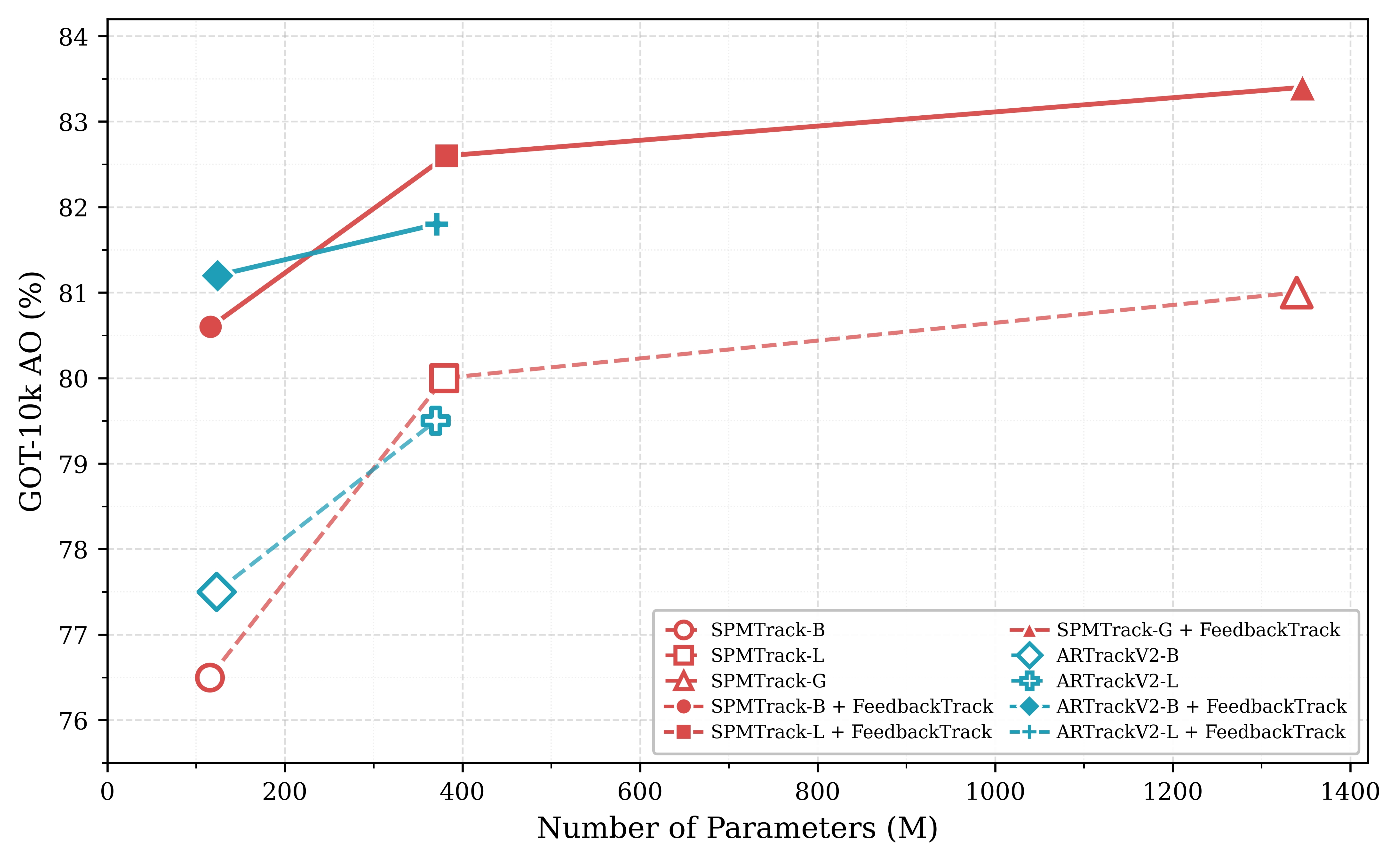}
    \caption{
        Performance versus model parameter count on GOT-10k for the
        SPMTrack and ARTrackV2 series.
        Solid and dashed lines denote the FeedbackTrack-enhanced and
        original models, respectively.
        FeedbackTrack consistently improves tracking performance across
        different backbone scales with only marginal parameter overhead.
    }
    \label{fig:performance_comparison}
\end{figure}

\begin{figure*}[t]
    \centering
    \includegraphics[width=\textwidth]{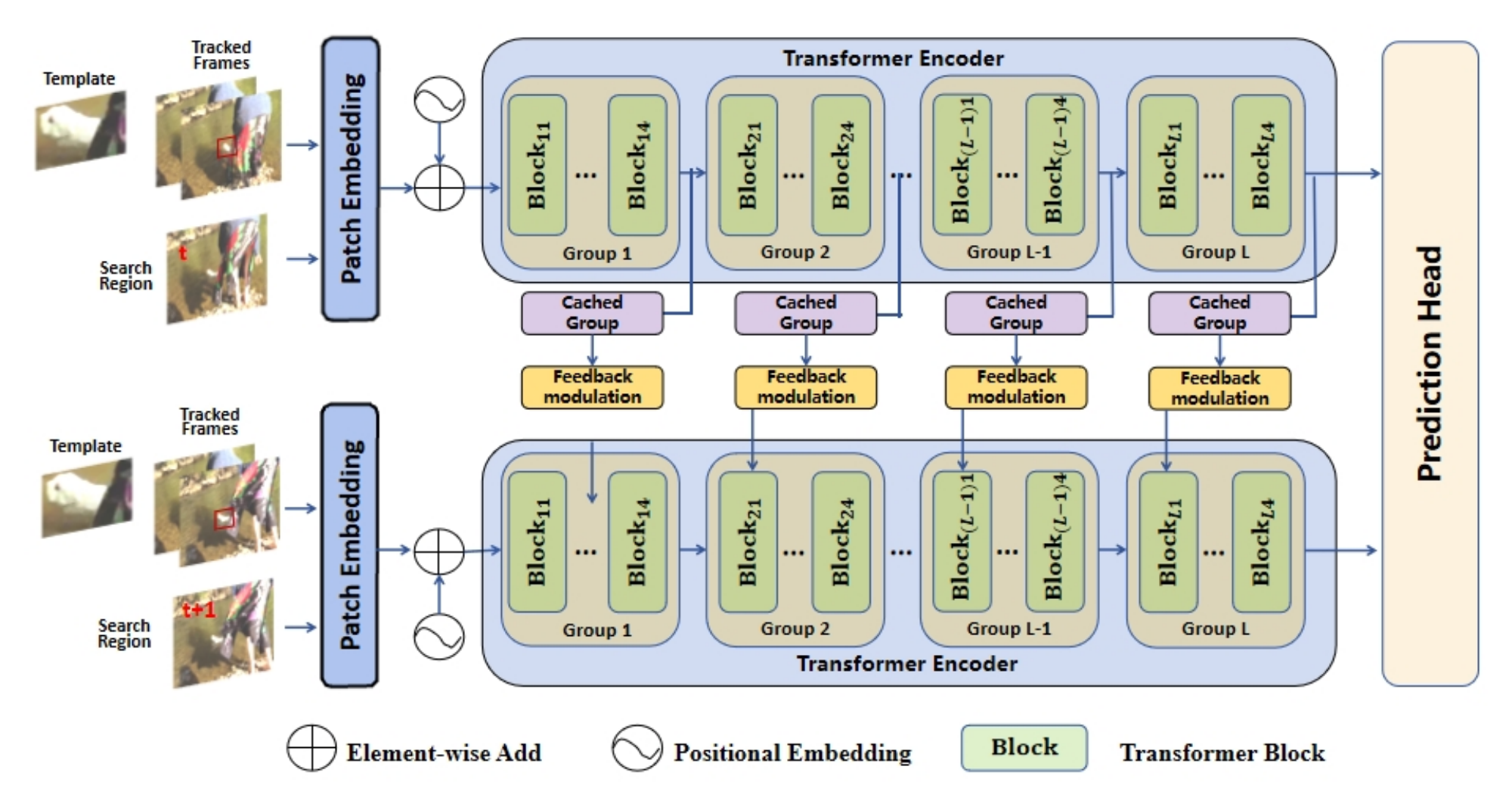}
    \caption{
        Overall architecture of FeedbackTrack.
        Previous-frame outputs of selected Transformer groups are detached,
        cached, and returned to the corresponding groups in the current frame.
        The original prediction head remains unchanged.
    }
    \label{fig:overall_structure}
\end{figure*}

\section{Introduction}
\label{sec:introduction}

Visual object tracking aims to continuously localize an arbitrary target
throughout a video given only its initial state.
A reliable tracker must preserve target identity while adapting to
appearance variation, occlusion, fast motion, background clutter, and
similar distractors.
Recent advances have shifted tracking architectures from convolutional
Siamese matching toward Transformer-based target--search interaction
\cite{siamrpnpp2019,transt2021,stark2021,mixformer2022}.
In particular, one-stream and sequence-based trackers jointly model
template and search information within a shared backbone, enabling strong
feature interaction and accurate target localization
\cite{ostrack2022,seqtrack2023,artrack2023,cai2025spmtrack}.

Despite their strong performance, existing Transformer trackers still
rely predominantly on feed-forward visual feature extraction within each
frame.
Temporal information has been incorporated through updated templates,
historical prompts, autoregressive queries, candidate associations, and
additional temporal tokens
\cite{wang2021transformertracker,keeptrack2021,hiptrack2024,
odtrack2024,aqatrack2024,artrackv22024}.
These mechanisms effectively preserve target history, but temporal states
are commonly introduced at the input, prompt, query, or prediction level
and then processed by another largely feed-forward visual encoder.
Intermediate representations formed at a particular backbone stage in
the previous frame are rarely returned to the corresponding processing
stage in the current frame.
Consequently, the visual feature hierarchy itself lacks an explicit
recurrent pathway through which previously formed representations can
directly modulate ongoing feature extraction.

Biological vision suggests a different computational organization.
The mammalian visual system combines ascending feed-forward pathways with
extensive recurrent and feedback connections, allowing previously formed
representations to influence subsequent sensory processing
\cite{lamme2000distinct,markov2014anatomy}.
Recent studies further indicate that visual feedback is distributed
across processing stages, pathway-specific, and modulatory rather than a
simple reversal of feed-forward activity
\cite{federer2021stream,siu2021direct,semedo2022feedforward,
fisek2023cortico,shen2022distinct}.
These findings motivate us to investigate whether recurrent reuse of
intermediate visual states can improve Transformer tracking.
We do not seek to reproduce specific cortical structures or neural
dynamics, but instead abstract the general principle that previously
formed representations can provide stage-corresponding modulation of
current visual processing.

To this end, we propose \textbf{FeedbackTrack}, a visual-cortex-inspired
framework that introduces sparse, group-level layer-aligned cross-frame
feedback into pretrained Transformer trackers.
The visual encoder is partitioned into groups of consecutive Transformer
blocks.
For each selected group, the previous-frame group output is detached,
cached, and returned to the first block of the corresponding group in the
current frame.
This design converts the originally stateless feature hierarchy into a
lightweight recurrent system while retaining the original feed-forward
pathway, tracking interface, and prediction modules.
Only one-frame states are retained, so the cache size remains constant
with respect to video length.

FeedbackTrack implements recurrent modulation through two complementary
pathways.
\textbf{Query Feedback} transforms previous search-token states through
a low-rank projection to generate token-wise biases for current search
queries.
A per-sample, per-head, and per-token RMS alignment operation controls the
bias magnitude relative to the pretrained queries.
\textbf{Gate Feedback} pools the complete previous group state and
generates bounded, context-dependent channel-group modulation of the
projected attention output.
Query Feedback therefore influences where current attention retrieves
evidence, whereas Gate Feedback regulates the resulting attention
response.

We instantiate FeedbackTrack on two representative tracking frameworks,
SPMTrack and ARTrackV2.
On SPMTrack, FeedbackTrack improves GOT-10k AO from 76.5 to 80.6 for
ViT-B, from 80.0 to 82.6 for ViT-L, and from 81.0 to 83.4 for ViT-G.
The corresponding LaSOT AUC scores increase from 74.9 to 76.4, from 76.8
to 77.9, and from 77.4 to 79.1.
FeedbackTrack also improves ARTrackV2-B and ARTrackV2-L to 81.2 and 81.8
AO on GOT-10k, with corresponding LaSOT AUC scores of 74.8 and 75.4.
Across all five configurations, the additional parameter cost remains
below 1\%.
Moreover, under the same modulation architecture and parameter count,
cross-frame feedback outperforms its same-frame counterpart by 3.2, 2.3,
and 1.8 AO points for ViT-B, ViT-L, and ViT-G, respectively, demonstrating
that the improvements primarily arise from recurrent historical
information rather than additional modulation capacity alone.

Although all Query Feedback scales are initialized identically, the
trained models develop a non-uniform depth-dependent organization, with
relatively weak feedback in shallow groups and stronger modulation in
intermediate and deep groups.
This emergent pattern is qualitatively consistent with the hierarchical
and pathway-dependent organization of biological visual feedback, while
representing a computational correspondence rather than an anatomical
equivalence.

Our main contributions are summarized as follows:

\begin{itemize}
    \item We introduce FeedbackTrack, a general framework that equips
    pretrained Transformer trackers with sparse, group-level layer-aligned
    cross-frame feedback, enabling intermediate previous-frame states to
    recurrently modulate corresponding stages of current-frame visual
    processing.

    \item We develop two complementary lightweight pathways.
    Query Feedback performs low-rank, RMS-aligned modulation of current
    search queries, while Gate Feedback applies bounded,
    context-dependent channel-group rescaling to the projected attention
    output.

    \item We demonstrate consistent improvements across two tracking
    frameworks and five backbone configurations on LaSOT and GOT-10k,
    while introducing less than 1\% additional parameters.
    Controlled same-frame comparisons further verify the importance of
    recurrent historical information.

    \item We analyze the learned feedback strengths and identify an
    emergent depth-dependent organization, showing that cross-frame
    feedback is allocated non-uniformly across visual representation
    stages.
\end{itemize}

\begin{figure}[t]
    \centering
    \includegraphics[width=\columnwidth]{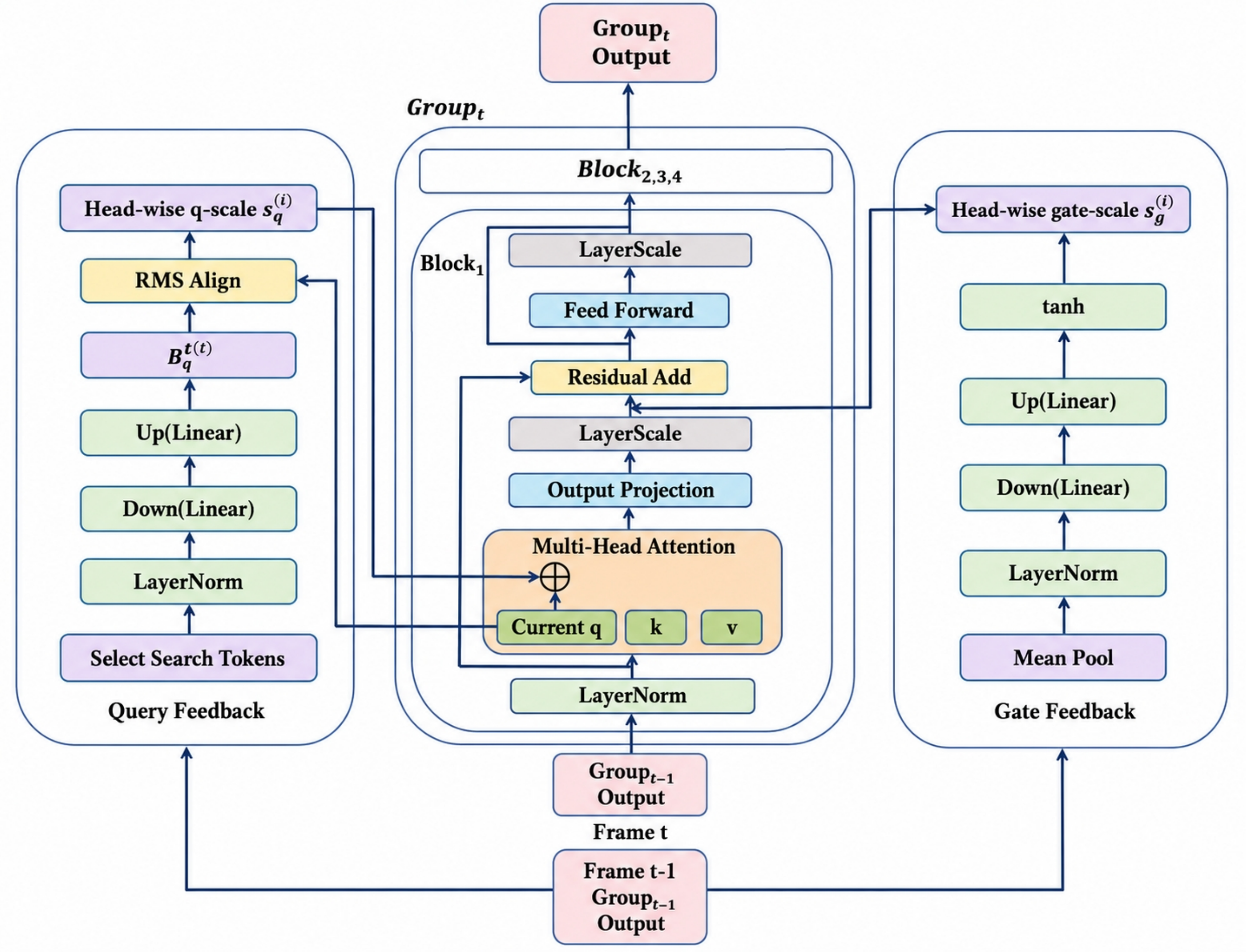}
    \caption{
        Feedback module.
        Query Feedback generates a norm-controlled bias for current
        search-token queries, while Gate Feedback modulates the projected
        attention output using pooled previous-frame context.
    }
    \label{fig:feedback_module}
\end{figure}

\section{Related Work}
\label{sec:related_work}

\subsection{Transformer-Based Visual Tracking}
\label{sec:transformer_tracking}

Visual object tracking has progressively shifted from convolutional
Siamese matching to Transformer-based architectures with explicit
target--search interaction. Representative methods include TransT,
STARK, OSTrack, SeqTrack, ARTrack, and ARTrackV2
\cite{transt2021,stark2021,ostrack2022,seqtrack2023,
artrack2023,artrackv22024}, while more recent approaches further explore
adaptive temporal queries and parameter-efficient tracking
\cite{aqatrack2024,cai2025spmtrack}. Although these trackers may use
historical templates, prompts, or autoregressive predictions,
intermediate visual representations are generally not recurrently
returned to corresponding backbone stages in the following frame.
FeedbackTrack complements these architectures by introducing sparse,
group-level layer-aligned cross-frame feedback inside the visual encoder.

\subsection{Temporal Modeling in Visual Tracking}
\label{sec:temporal_tracking}

Existing trackers exploit temporal context through historical templates,
candidate associations, memory representations, prompts, autoregressive
queries, and temporal tokens
\cite{wang2021transformertracker,keeptrack2021,hiptrack2024,
odtrack2024,aqatrack2024,artrackv22024}. These approaches typically
introduce temporal information at the input, prompt, memory, or prediction
level, after which the visual encoder remains predominantly feed-forward.
In contrast, FeedbackTrack treats intermediate Transformer group outputs
as recurrent visual states and returns each previous-frame group output
to the corresponding current-frame group. This design is complementary
to existing temporal modeling strategies, including the prompt-based
autoregressive pipeline of ARTrackV2.

\subsection{Recurrent Feedback in Biological Vision}
\label{sec:biological_feedback}

Biological vision combines feed-forward and recurrent processing, with
distributed and pathway-specific feedback modulating ongoing sensory
representations across the visual hierarchy
\cite{lamme2000distinct,markov2014anatomy,
federer2021stream,semedo2022feedforward,
fisek2023cortico,shen2022distinct}. FeedbackTrack does not reproduce
specific cortical structures or neural pathways, but abstracts several
general computational principles, including recurrent reuse of previous
representations, correspondence between feedback states and processing
stages, and residual modulation of current visual computation. These
principles are implemented through group-level layer-aligned cross-frame
connections together with Query Feedback and Gate Feedback.

\section{Method}
\label{sec:method}

\subsection{Overview}
\label{sec:method_overview}

Biological visual processing combines feed-forward computation with
recurrent feedback that modulates ongoing sensory representations
\cite{federer2021stream,siu2021direct,semedo2022feedforward,
fisek2023cortico,shen2022distinct}.
Inspired by this computational principle, FeedbackTrack introduces
cross-frame recurrent modulation into the visual Transformer encoder,
without attempting to reproduce specific cortical structures.

FeedbackTrack is formulated as a generic feedback mechanism for
Transformer-based trackers and is instantiated on two representative
frameworks, SPMTrack~\cite{cai2025spmtrack} and
ARTrackV2~\cite{artrackv22024}.
For SPMTrack, the feedback modules are inserted into its DINOv2-based
Transformer backbone
\cite{oquab2024dinov2,dosovitskiy2021image},
while the patch embedding, tracking embeddings, TMoE adapters, and
prediction head remain unchanged.
For ARTrackV2, only the visual encoder is modified, while the original
trajectory prompts, appearance prompts, autoregressive prediction
modules, and training objective are retained.
As illustrated in Fig.~\ref{fig:overall_structure}, FeedbackTrack
augments the original visual feature extractor with sparse cross-frame
connections while preserving the task-specific tracking interface and
prediction pipeline.

\subsection{Layer-Aligned Cross-Frame Feedback}
\label{sec:layer_aligned_feedback}

We partition the Transformer backbone into groups of $G$ consecutive
blocks.
Let $\mathbf{X}_{t}^{(i)}$ and $\mathbf{U}_{t}^{(i)}$ denote the input
and output of group $i$ at frame $t$, respectively.
The output of the corresponding group in the previous frame is detached
and stored as a one-frame cache:

\begin{equation}
    \mathbf{C}_{t-1}^{(i)}
    =
    \operatorname{StopGrad}
    \left(
        \mathbf{U}_{t-1}^{(i)}
    \right).
    \label{eq:feedback_cache}
\end{equation}

At frame $t$, the cached representation is provided to the first
Transformer block of the same group:

\begin{equation}
    \mathbf{U}_{t}^{(i)}
    =
    \mathcal{G}_{\mathrm{fb}}^{(i)}
    \left(
        \mathbf{X}_{t}^{(i)},
        \mathbf{C}_{t-1}^{(i)}
    \right),
    \label{eq:feedback_group}
\end{equation}

where $\mathcal{G}_{\mathrm{fb}}^{(i)}$ denotes group $i$ equipped with
the proposed feedback module.

Because the output of group $i$ at frame $t-1$ is returned to the first
block of the same group $i$ at frame $t$, the temporal connection
preserves the hierarchical stage at which the representation was
generated.
We therefore refer to this design as
\emph{group-level layer-aligned cross-frame feedback}.
Only the first block of each group contains the feedback module, whereas
the remaining blocks retain their original computation.

We use a group size of $G=4$.
For example, the output produced after blocks $0$--$3$ at frame $t-1$
is returned to block $0$ at frame $t$, while the output produced after
blocks $4$--$7$ is returned to block $4$.
Accordingly, feedback modules are inserted at blocks
$\{0,4,8\}$ for ViT-B,
$\{0,4,8,12,16,20\}$ for ViT-L, and
$\{0,4,8,12,16,20,24,28,32,36\}$ for ViT-G.
The same grouping strategy is applied to the visual encoders of
ARTrackV2-B and ARTrackV2-L.

\subsection{Feedback Module}
\label{sec:feedback_module}

The feedback module contains two complementary pathways, as illustrated
in Fig.~\ref{fig:feedback_module}.
Query Feedback preserves token-level spatial information and recurrently
modulates the current search-token queries.
Gate Feedback summarizes the complete previous group state and applies
context-dependent channel-group modulation to the projected attention
output.
The two pathways therefore inject historical information at different
locations in the attention computation.

\subsubsection{Norm-Controlled Query Feedback}
\label{sec:query_feedback}

Let
\begin{equation}
    \mathbf{M}_{t-1}^{q,(i)}
    \in
    \mathbb{R}^{B\times N_s\times D}
\end{equation}
denote the search-token subset selected from the cached representation
$\mathbf{C}_{t-1}^{(i)}$, where $B$ is the batch size, $N_s$ is the
number of search tokens, and $D$ is the feature dimension.
We first transform the historical search-token states using LayerNorm
and a low-rank projection:

\begin{equation}
    \mathbf{B}_{t}^{q,(i)}
    =
    \operatorname{Up}^{(i)}
    \left(
        \operatorname{Down}^{(i)}
        \left(
            \operatorname{LN}
            \left(
                \mathbf{M}_{t-1}^{q,(i)}
            \right)
        \right)
    \right),
    \label{eq:query_projection}
\end{equation}

where the down- and up-projections map the feature dimension as
$D\rightarrow r\rightarrow D$, with a bottleneck dimension of $r=16$.
The projected historical bias is then reshaped into the multi-head form

\begin{equation}
    \mathbf{B}_{t}^{q,(i)}
    \in
    \mathbb{R}^{B\times H\times N_s\times d_h},
\end{equation}

where $H$ is the number of attention heads and $D=Hd_h$.

Directly adding the projected historical bias to the pretrained queries
may result in a substantial magnitude mismatch.
We therefore align their root-mean-square magnitudes independently for
each sample, attention head, and search token:

\begin{equation}
    \widehat{\mathbf{B}}_{t,bhs:}^{q,(i)}
    =
    \mathbf{B}_{t,bhs:}^{q,(i)}
    \frac{
        \rho\,
        \operatorname{RMS}
        \left(
            \mathbf{Q}_{t,bhs:}^{(i)}
        \right)
    }{
        \operatorname{RMS}
        \left(
            \mathbf{B}_{t,bhs:}^{q,(i)}
        \right)
    },
    \label{eq:rms_alignment}
\end{equation}

where $b$, $h$, and $s$ index the sample, attention head, and search
token, respectively.
The RMS operator is defined as

\begin{equation}
    \operatorname{RMS}(\mathbf{x})
    =
    \sqrt{
        \frac{1}{d_h}
        \sum_{d=1}^{d_h}x_d^2
        +
        \epsilon
    }.
    \label{eq:rms_definition}
\end{equation}

We set $\rho=0.05$ and $\epsilon=10^{-4}$.
The aligned historical bias is added to the current search-token queries
using a learnable head-wise scale vector
$\mathbf{s}_{q}^{(i)}\in\mathbb{R}^{H}$:

\begin{equation}
    \widetilde{\mathbf{Q}}_{t,bhs:}^{(i)}
    =
    \mathbf{Q}_{t,bhs:}^{(i)}
    +
    s_{q,h}^{(i)}
    \widehat{\mathbf{B}}_{t,bhs:}^{q,(i)}.
    \label{eq:query_modulation}
\end{equation}

All $q$-scales are initialized to $0.01$, allowing the modified tracker
to start close to the pretrained model.
Only the current search-token queries are modified, while all non-search
queries, keys, and values remain unchanged.

\subsubsection{Gate Feedback}
\label{sec:gate_feedback}

Query Feedback is complemented by an output-side modulation pathway.
We first summarize the complete previous group state by averaging over
all cached tokens:

\begin{equation}
    \mathbf{p}_{t-1}^{(i)}
    =
    \frac{1}{N}
    \sum_{n=1}^{N}
    \mathbf{C}_{t-1,n}^{(i)},
    \label{eq:gate_context}
\end{equation}

where
$\mathbf{p}_{t-1}^{(i)}\in\mathbb{R}^{B\times D}$
and $N$ denotes the total number of cached tokens.

A lightweight MLP maps the pooled context from $D$ to a hidden
dimension of $384$ and then to $H$ channel groups.
A tanh-bounded contextual signal is combined with a learnable
channel-group scale vector
$\mathbf{s}_{g}^{(i)}\in\mathbb{R}^{H}$:

\begin{equation}
    \mathbf{g}_{t}^{(i)}
    =
    \tanh
    \left(
        \operatorname{MLP}_{g}^{(i)}
        \left(
            \operatorname{LN}
            \left(
                \mathbf{p}_{t-1}^{(i)}
            \right)
        \right)
    \right)
    \odot
    \mathbf{s}_{g}^{(i)},
    \label{eq:gate_generation}
\end{equation}

where
$\mathbf{g}_{t}^{(i)}\in\mathbb{R}^{B\times H}$
contains one modulation coefficient for each channel group.
The learnable gate scales are initialized to $0.01$.

Let $\mathbf{A}_{t}^{(i)}$ denote the attention output after head
concatenation, output projection, projection dropout, and the original
LayerScale.
We reshape its feature dimension into $H$ channel groups, each containing
$d_h$ channels, and apply

\begin{equation}
    \widetilde{\mathbf{A}}_{t,bnh:}^{(i)}
    =
    \left(
        1+g_{t,bh}^{(i)}
    \right)
    \mathbf{A}_{t,bnh:}^{(i)}.
    \label{eq:gate_modulation}
\end{equation}

The same gate is broadcast across all tokens and applied before the
original residual connection.
Because the output projection has already mixed information from the
original attention heads, this operation should be interpreted as
channel-group rescaling of the projected attention output rather than
direct modulation of individual attention heads.

\subsection{Training and Online Inference}
\label{sec:training_inference}

During training, consecutive search frames are processed sequentially,
and detached previous-frame group outputs serve as feedback caches for
the current frame, avoiding cross-frame backpropagation through time.
The original feed-forward and residual pathways are retained, while the
small initial query and gate scales keep the model close to the pretrained
base tracker.

During inference, each selected Transformer group maintains a one-frame
cache that is updated after every frame.
When no valid cache is available, the original base-tracker computation is
used.
Because only the immediately preceding group states are stored, the cache
memory remains constant with respect to video length.

\begin{table*}[!t]
    \centering
    \caption{
        Comparison with state-of-the-art trackers on LaSOT and
        GOT-10k.
        Each indented FeedbackTrack row is built upon the
        immediately preceding baseline.
        The best and second-best results are highlighted in bold
        and underlined, respectively.
    }
    \label{tab:main_results}
    \footnotesize
    \setlength{\tabcolsep}{3.8pt}
    \renewcommand{\arraystretch}{0.92}
    \resizebox{0.94\textwidth}{!}{
    \begin{tabular}{lccccccc}
        \toprule
        \multirow{2}{*}{Method}
        & \multirow{2}{*}{Source}
        & \multicolumn{3}{c}{LaSOT}
        & \multicolumn{3}{c}{GOT-10k} \\
        \cmidrule(lr){3-5}
        \cmidrule(lr){6-8}
        &
        & AUC
        & $P_{\mathrm{Norm}}$
        & $P$
        & AO
        & $\mathrm{SR}_{0.5}$
        & $\mathrm{SR}_{0.75}$ \\
        \midrule

        AiATrack
        & ECCV'22
        & 69.0 & 79.4 & 73.8
        & 69.6 & 80.0 & 63.2 \\

        OSTrack$_{384}$
        & ECCV'22
        & 71.1 & 81.1 & 77.6
        & 73.7 & 83.2 & 70.8 \\

        ARTrack$_{384}$
        & CVPR'23
        & 72.6 & 81.7 & 79.1
        & 75.5 & 84.3 & 74.3 \\

        GRM
        & CVPR'23
        & 69.9 & 79.3 & 75.8
        & 73.4 & 82.9 & 70.4 \\

        SeqTrack-B$_{384}$
        & CVPR'23
        & 71.5 & 81.1 & 77.8
        & 74.5 & 84.3 & 71.4 \\

        F-BDMTrack$_{384}$
        & ICCV'23
        & 72.0 & 81.5 & 77.7
        & 75.4 & 84.3 & 72.9 \\

        ROMTrack$_{384}$
        & ICCV'23
        & 71.4 & 81.4 & 78.2
        & 74.2 & 84.3 & 72.4 \\

        ODTrack-B
        & AAAI'24
        & 73.2 & 83.2 & 80.6
        & 77.0 & 87.9 & 75.1 \\

        AQATrack$_{384}$
        & CVPR'24
        & 72.7 & 82.9 & 80.2
        & 76.0 & 85.2 & 74.9 \\

        ARTrackV2-B$_{384}$
        & CVPR'24
        & 73.0 & 82.0 & 79.6
        & 77.5 & 86.0 & 75.5 \\

        \quad + FeedbackTrack
        & Ours
        & 74.8 & 82.7 & 80.2
        & 81.2 & 89.2 & 81.3 \\

        ARTrackV2-L$_{384}$
        & CVPR'24
        & 73.6 & 82.8 & 81.1
        & 79.5 & 87.8 & 79.6 \\

        \quad + FeedbackTrack
        & Ours
        & 75.4 & 83.6 & 81.6
        & 81.8 & 89.6 & 82.1 \\

        HIPTrack
        & CVPR'24
        & 72.7 & 82.9 & 79.5
        & 77.4 & 88.0 & 74.5 \\

        LoRAT-B$_{378}$
        & ECCV'24
        & 72.9 & 81.9 & 79.1
        & 73.7 & 82.6 & 72.9 \\

        LoRAT-L$_{378}$
        & ECCV'24
        & 75.1 & 84.1 & 82.0
        & 77.5 & 86.2 & 78.1 \\

        LoRAT-G$_{378}$
        & ECCV'24
        & 76.2 & 85.3 & 83.5
        & 78.9 & 87.8 & 80.7 \\

        SPMTrack-B
        & CVPR'25
        & 74.9 & 84.0 & 81.7
        & 76.5 & 85.9 & 76.3 \\

        \quad + FeedbackTrack
        & Ours
        & 76.4 & 84.9 & 82.6
        & 80.6 & 89.9 & 80.4 \\

        SPMTrack-L
        & CVPR'25
        & 76.8 & 85.9 & 84.0
        & 80.0 & 89.4 & 79.9 \\

        \quad + FeedbackTrack
        & Ours
        & \underline{77.9}
        & \underline{86.8}
        & \underline{85.2}
        & \underline{82.6}
        & \underline{91.4}
        & \underline{83.8} \\

        SPMTrack-G
        & CVPR'25
        & 77.4 & 86.6 & 85.0
        & 81.0 & 89.2 & 82.3 \\

        \quad + FeedbackTrack
        & Ours
        & \textbf{79.1}
        & \textbf{87.8}
        & \textbf{86.3}
        & \textbf{83.4}
        & \textbf{91.8}
        & \textbf{84.0} \\

        \bottomrule
    \end{tabular}
    }
\end{table*}

\section{Experiments}
\label{sec:experiments}

\subsection{Datasets and Evaluation Metrics}

We evaluate FeedbackTrack on two widely used visual tracking
benchmarks, LaSOT~\cite{fan2019lasot} and
GOT-10k~\cite{huang2019got10k}.
For LaSOT, we report the area under the success curve (AUC),
normalized precision ($P_{\mathrm{Norm}}$), and precision ($P$).
For GOT-10k, we report average overlap (AO) and success rates
at overlap thresholds of 0.5 and 0.75, denoted as
$\mathrm{SR}_{0.5}$ and $\mathrm{SR}_{0.75}$, respectively.

\subsection{Implementation Details}

We evaluate FeedbackTrack on two representative Transformer tracking
frameworks, SPMTrack~\cite{cai2025spmtrack} and
ARTrackV2~\cite{artrackv22024}.
For SPMTrack, FeedbackTrack is implemented in TrackIt and evaluated with
ViT-B, ViT-L, and ViT-G backbones.
For ARTrackV2-B and ARTrackV2-L, the feedback modules are inserted only
into the visual encoder, while the original prompts, autoregressive
prediction modules, and training objectives remain unchanged.

The SPMTrack-based models are trained with AdamW on four
NVIDIA A100-SXM4-80GB GPUs using a total batch size of 32.
The base learning rate is $1\times10^{-5}$ with a weight decay of 0.1.
We use a per-iteration cosine schedule with a 10-epoch warm-up from
$1\times10^{-7}$ and a minimum learning rate of $1\times10^{-6}$.
Most backbone parameters are frozen, while the last four Transformer
blocks are fine-tuned at $1\times10^{-6}$.
The TMoE adapters, tracking embeddings, and prediction head use the base
learning rate, whereas Query Feedback and Gate Feedback are optimized at
$5\times10^{-5}$.
For ARTrackV2, we follow its original training configuration.

Inference throughput is measured on a single NVIDIA A100 GPU with batch
size 1.
FPS is averaged over all evaluated GOT-10k test frames, excluding image
loading and preprocessing.


\subsection{Main Results}

Table~\ref{tab:main_results} compares FeedbackTrack with
representative state-of-the-art trackers on LaSOT and GOT-10k.
When applied to SPMTrack, FeedbackTrack consistently improves all six
metrics across the B, L, and G variants.
On GOT-10k, the AO scores increase by 4.1, 2.6, and 2.4 points,
respectively, while the corresponding LaSOT AUC gains are 1.5, 1.1,
and 1.7 points.

FeedbackTrack also generalizes well to ARTrackV2.
It improves ARTrackV2-B and ARTrackV2-L to 81.2 and 81.8 AO on
GOT-10k, with LaSOT AUC scores of 74.8 and 75.4.
The corresponding $\mathrm{SR}_{0.75}$ gains are 5.8 and 2.5 points,
showing that intermediate visual-state feedback complements the
autoregressive temporal modeling of ARTrackV2.

\subsection{Ablation Study}


\begin{table}[!t]
    \centering
    \caption{
        Ablation of Query Feedback (QF) and Gate Feedback (GF)
        on GOT-10k.
    }
    \label{tab:component_ablation}
    \small
    \setlength{\tabcolsep}{4.0pt}
    \begin{tabular}{lccccc}
        \toprule
        Backbone
        & QF
        & GF
        & AO
        & $\mathrm{SR}_{0.5}$
        & $\mathrm{SR}_{0.75}$ \\
        \midrule

        ViT-B
        &
        &
        & 76.5 & 85.9 & 76.3 \\

        ViT-B
        & $\checkmark$
        &
        & 79.7 & 89.2 & 80.1 \\

        ViT-B
        &
        & $\checkmark$
        & 78.3 & 87.2 & 77.8 \\

        ViT-B
        & $\checkmark$
        & $\checkmark$
        & \textbf{80.6}
        & \textbf{89.9}
        & \textbf{80.4} \\

        \midrule

        ViT-L
        &
        &
        & 80.0 & 89.4 & 79.9 \\

        ViT-L
        & $\checkmark$
        &
        & 82.0 & 91.1 & 83.4 \\

        ViT-L
        &
        & $\checkmark$
        & 80.6 & 90.0 & 81.2 \\

        ViT-L
        & $\checkmark$
        & $\checkmark$
        & \textbf{82.6}
        & \textbf{91.4}
        & \textbf{83.8} \\

        \midrule

        ViT-G
        &
        &
        & 81.0 & 89.2 & 82.3 \\

        ViT-G
        & $\checkmark$
        &
        & 83.0 & 91.4 & 83.7 \\

        ViT-G
        &
        & $\checkmark$
        & 81.8 & 90.2 & 82.7 \\

        ViT-G
        & $\checkmark$
        & $\checkmark$
        & \textbf{83.4}
        & \textbf{91.8}
        & \textbf{84.0} \\

        \bottomrule
    \end{tabular}
\end{table}

\paragraph{Effect of feedback components.}

Table~\ref{tab:component_ablation} shows that both Query Feedback and
Gate Feedback are independently effective.
Query Feedback contributes the majority of the gain, improving AO by
3.2, 2.0, and 2.0 points for ViT-B, ViT-L, and ViT-G, while Gate
Feedback yields gains of 1.8, 0.6, and 0.8 points, respectively.
Combining both pathways achieves the best results on all metrics,
indicating that token-level query modulation and global context-based
output modulation are complementary.

\begin{figure*}[!t]
    \centering
    \begin{tabular}{ccc}
        \includegraphics[width=0.31\textwidth]{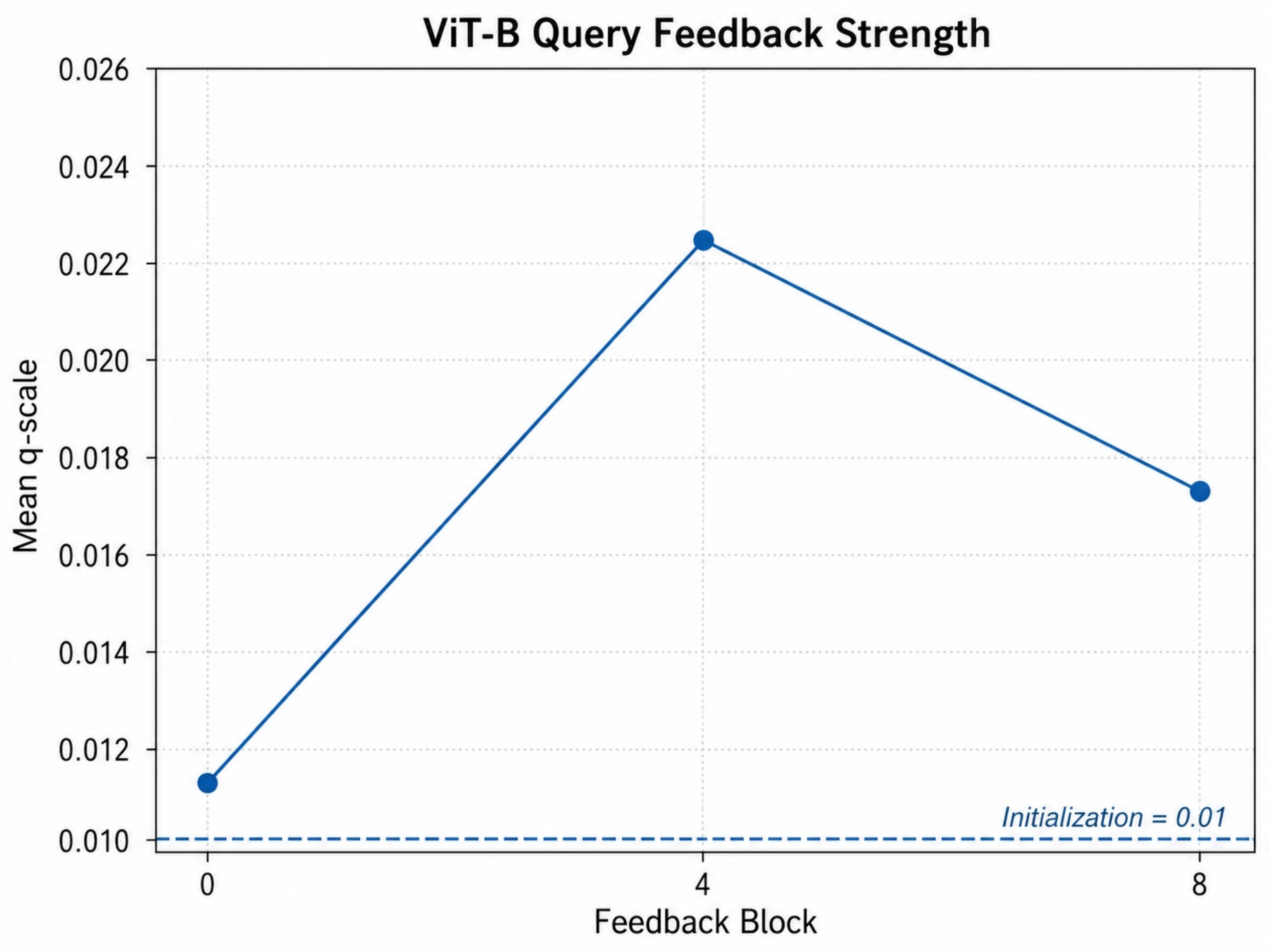}
        &
        \includegraphics[width=0.31\textwidth]{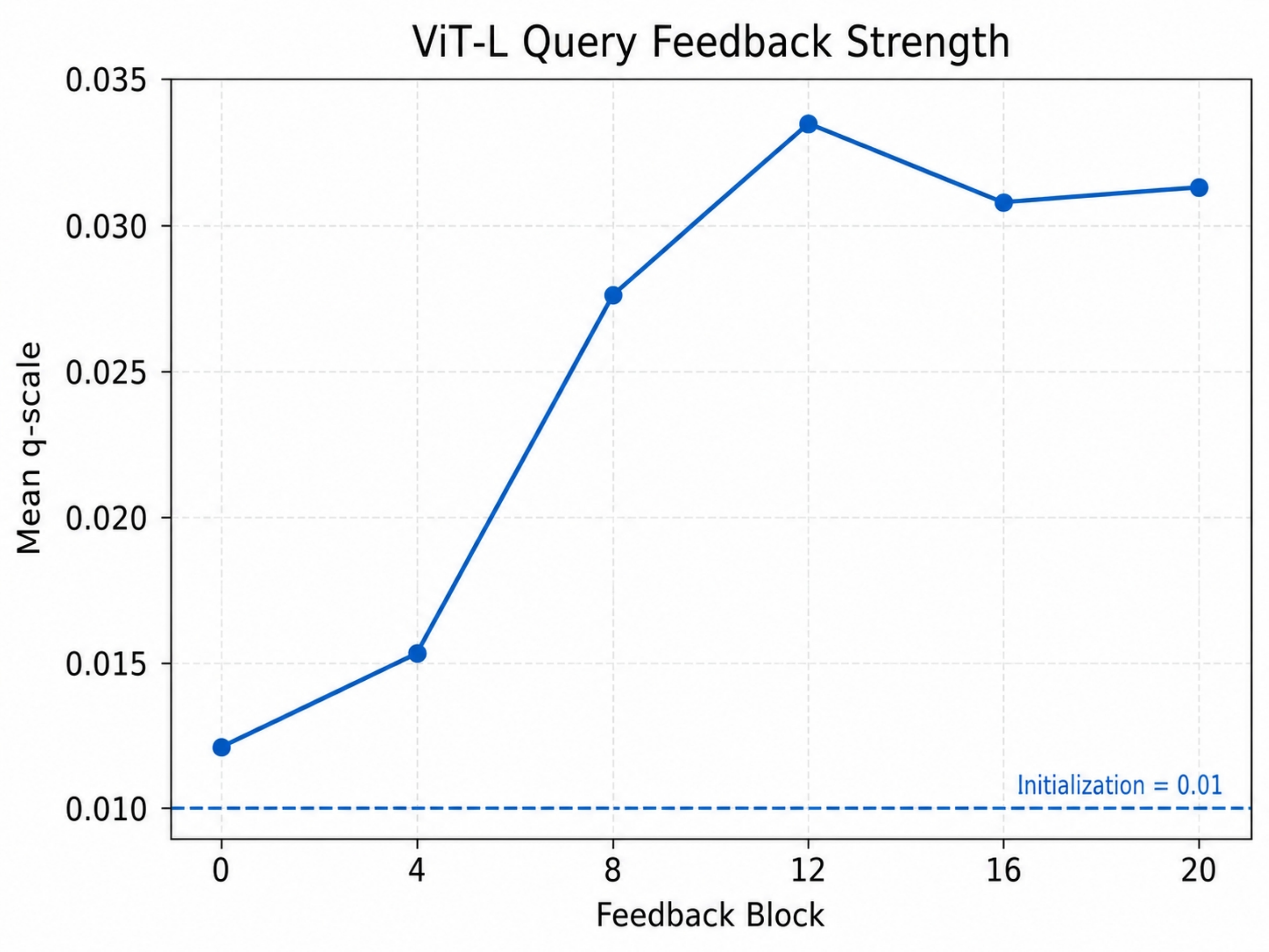}
        &
        \includegraphics[width=0.31\textwidth]{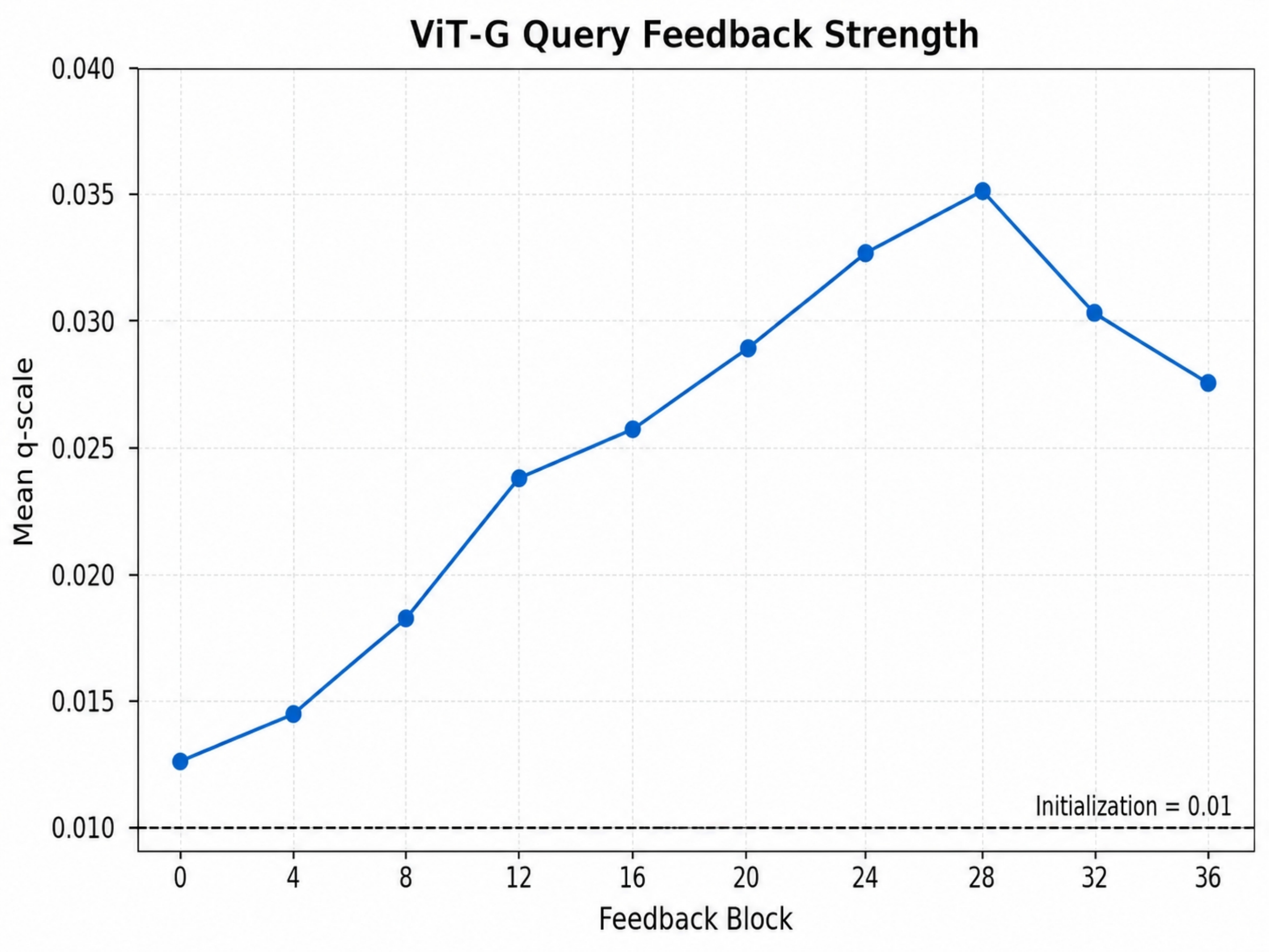}
        \\
        (a) ViT-B
        &
        (b) ViT-L
        &
        (c) ViT-G
    \end{tabular}
    \caption{
        Learned signed mean $q$-scale across feedback blocks,
        averaged over attention heads within each block.
        All $q$-scales are initialized to 0.01, as indicated by
        the dashed lines.
    }
    \label{fig:qscale_analysis}
\end{figure*}

\begin{table}[!t]
    \centering
    \caption{
        Comparison between same-frame modulation and
        cross-frame feedback on GOT-10k.
        Same-frame modulation uses the detached current group
        input, while cross-frame feedback uses the detached
        previous-frame group output.
        Both variants use the same modules, insertion positions,
        and parameter count.
    }
    \label{tab:cross_frame_ablation}
    \small
    \setlength{\tabcolsep}{4.0pt}
    \begin{tabular}{llccc}
        \toprule
        Backbone
        & Modulation Source
        & AO
        & $\mathrm{SR}_{0.5}$
        & $\mathrm{SR}_{0.75}$ \\
        \midrule

        ViT-B
        & None
        & 76.5 & 85.9 & 76.3 \\

        ViT-B
        & Current-frame input
        & 77.4 & 86.3 & 77.7 \\

        ViT-B
        & Previous-frame output
        & \textbf{80.6}
        & \textbf{89.9}
        & \textbf{80.4} \\

        \midrule

        ViT-L
        & None
        & 80.0 & 89.4 & 79.9 \\

        ViT-L
        & Current-frame input
        & 80.3 & 90.2 & 81.1 \\

        ViT-L
        & Previous-frame output
        & \textbf{82.6}
        & \textbf{91.4}
        & \textbf{83.8} \\

        \midrule

        ViT-G
        & None
        & 81.0 & 89.2 & 82.3 \\

        ViT-G
        & Current-frame input
        & 81.6 & 89.7 & 83.0 \\

        ViT-G
        & Previous-frame output
        & \textbf{83.4}
        & \textbf{91.8}
        & \textbf{84.0} \\

        \bottomrule
    \end{tabular}
\end{table}

\paragraph{Cross-frame feedback versus same-frame modulation.}

To isolate the effect of recurrent historical information, we construct
a same-frame control that replaces the detached previous-frame group
output with the detached current group input, while keeping the same
modules, insertion positions, and parameter count.
As shown in Table~\ref{tab:cross_frame_ablation}, cross-frame feedback
outperforms this control by 3.2, 2.3, and 1.8 AO points for ViT-B,
ViT-L, and ViT-G, respectively, confirming that the main gains arise
from recurrent historical information rather than additional modulation
capacity.


\begin{table}[!t]
    \centering
    \caption{
        Effect of RMS alignment on GOT-10k.
        All variants contain both Query Feedback and Gate
        Feedback.
    }
    \label{tab:rms_alignment}
    \small
    \setlength{\tabcolsep}{5.0pt}
    \begin{tabular}{lcccc}
        \toprule
        Backbone
        & RMS Align.
        & AO
        & $\mathrm{SR}_{0.5}$
        & $\mathrm{SR}_{0.75}$ \\
        \midrule

        ViT-B
        &
        & 78.2 & 88.6 & 79.3 \\

        ViT-B
        & $\checkmark$
        & \textbf{80.6}
        & \textbf{89.9}
        & \textbf{80.4} \\

        \midrule

        ViT-L
        &
        & 82.1 & 90.3 & 82.6 \\

        ViT-L
        & $\checkmark$
        & \textbf{82.6}
        & \textbf{91.4}
        & \textbf{83.8} \\

        \midrule

        ViT-G
        &
        & 82.1 & 90.3 & 83.8 \\

        ViT-G
        & $\checkmark$
        & \textbf{83.4}
        & \textbf{91.8}
        & \textbf{84.0} \\

        \bottomrule
    \end{tabular}
\end{table}

\paragraph{Effect of RMS alignment.}

Table~\ref{tab:rms_alignment} evaluates RMS alignment in Query Feedback.
Aligning the historical bias with the magnitude of current search queries
improves all metrics, increasing AO by 2.4, 0.5, and 1.3 points for
ViT-B, ViT-L, and ViT-G, respectively.
This confirms the importance of controlling feedback magnitude when
modulating pretrained query representations.

\subsection{Analysis}


\begin{table}[!t]
    \centering
    \caption{
        Parameter overhead of FeedbackTrack across SPMTrack
        and ARTrackV2.
    }
    \label{tab:parameter_efficiency}
    \small
    \setlength{\tabcolsep}{3.0pt}
    \resizebox{\columnwidth}{!}{
    \begin{tabular}{lccccc}
        \toprule
        Model
        & Baseline
        & QF Only
        & QF + GF
        & Extra
        & Increase \\
        \midrule

        SPMTrack-B
        & 115.331M
        & 115.409M
        & 116.313M
        & 0.983M
        & 0.852\% \\

        SPMTrack-L
        & 379.582M
        & 379.791M
        & 382.202M
        & 2.620M
        & 0.690\% \\

        SPMTrack-G
        & 1339.512M
        & 1340.034M
        & 1346.060M
        & 6.548M
        & 0.489\% \\

        ARTrackV2-B$_{384}$
        & 130.514M
        & 130.593M
        & 131.497M
        & 0.983M
        & 0.753\% \\

        ARTrackV2-L$_{384}$
        & 382.059M
        & 382.268M
        & 384.679M
        & 2.620M
        & 0.686\% \\

        \bottomrule
    \end{tabular}
    }
\end{table}

\paragraph{Parameter efficiency.}

Table~\ref{tab:parameter_efficiency} reports the parameter
overhead across both tracking frameworks.
For the SPMTrack-based implementations, FeedbackTrack adds
0.983M, 2.620M, and 6.548M parameters to the B, L, and G
variants, corresponding to increases of only 0.852\%, 0.690\%,
and 0.489\%, respectively.
For ARTrackV2-B and ARTrackV2-L, the corresponding increases
are 0.983M and 2.620M, accounting for only 0.753\% and
0.686\% of the original model sizes.
FeedbackTrack therefore introduces less than 1\% parameter
overhead across all five model configurations.


\begin{table}[!t]
    \centering
    \caption{
        Model inference throughput on the GOT-10k test set.
        Image loading and preprocessing are excluded.
    }
    \label{tab:runtime_efficiency}
    \small
    \setlength{\tabcolsep}{4.2pt}
    \begin{tabular}{lccc}
        \toprule
        Model
        & Baseline FPS
        & FeedbackTrack FPS
        & Reduction \\
        \midrule

        SPMTrack-B
        & 45.59
        & 42.20
        & 7.44\% \\

        SPMTrack-L
        & 19.98
        & 18.82
        & 5.81\% \\

        SPMTrack-G
        & 4.40
        & 4.16
        & 5.45\% \\

        ARTrackV2-B$_{384}$
        & 15.7785
        & 15.7505
        & 0.18\% \\

        ARTrackV2-L$_{384}$
        & 12.9309
        & 12.8203
        & 0.86\% \\

        \bottomrule
    \end{tabular}
\end{table}

\paragraph{Runtime efficiency.}

Table~\ref{tab:runtime_efficiency} reports model inference
throughput.
For SPMTrack-B, SPMTrack-L, and SPMTrack-G, the throughput
reductions are 7.44\%, 5.81\%, and 5.45\%, respectively.
For ARTrackV2-B and ARTrackV2-L, the reductions are only
0.18\% and 0.86\%.
These results show that FeedbackTrack preserves most of the
original inference throughput across both tracking frameworks
while introducing explicit recurrent cross-frame modulation.


\paragraph{Biologically inspired analysis.}

Biological visual feedback is distributed non-uniformly across the
cortical hierarchy
\cite{federer2021stream,semedo2022feedforward,
fisek2023cortico,shen2022distinct}.
We therefore analyze the signed mean $q$-scale across attention heads
within each feedback block.
Since all scales are initialized to 0.01, their depth-dependent variation
is learned during optimization.

As shown in Fig.~\ref{fig:qscale_analysis}, ViT-B reaches its strongest
feedback in the intermediate group, ViT-L develops stronger modulation
from middle to deep groups, and ViT-G progressively increases toward deep
groups before slightly decreasing.
Overall, feedback is weaker in shallow groups and stronger in
intermediate and deep groups.
This pattern is qualitatively consistent with hierarchical biological
feedback, but reflects computational correspondence rather than
anatomical equivalence.

\section{Conclusion}

We presented FeedbackTrack, a visual-cortex-inspired framework that introduces sparse, group-level layer-aligned cross-frame feedback into Transformer trackers. Through complementary Query Feedback and Gate Feedback pathways, previous-frame representations recurrently modulate current visual processing while retaining the original feed-forward pathway and tracking pipeline. Beyond visual tracking, the proposed mechanism provides a general approach for introducing recurrent temporal states into pretrained video models. Future work will explore its application to broader video tasks, including video object segmentation, action recognition, video understanding, and video generation.

\bibliography{aaai2027}


\end{document}